\documentclass[sigconf]{acmart}   
\AtBeginDocument{%
  }

\setcopyright{none}

\acmConference{}
\acmISBN{}
\acmDOI{}
\acmYear{}
\copyrightyear{}

\acmISBN{}
\usepackage{multirow}

\begin{document}

\title{Text-to-Image Generation for Projector-Camera System Registration}


\author{Xinyu Chen}
\authornote{First author.}
\email{2211100247@nbu.edu.cn}
\affiliation{%
  \institution{Ningbo University}
  \city{Ningbo}
  \country{China}
}

\author{Yuqi Li}
\authornotemark[1]
\authornote{Corresponding author.}
\authornotemark[0]
\email{liyuqi1@nbu.edu.cn}
\affiliation{%
  \institution{Ningbo University}
  \city{Ningbo}
  \country{China}}

\author{Jiabao Li}
\email{jiabaoL6@uci.edu}
\affiliation{%
  \institution{University of California, Irvine}
  \country{United States}
}

\author{Pinyan Tang}
\email{tangpinyan@nbu.edu.cn}
\affiliation{%
  \institution{Ningbo University}
  \city{Ningbo}
  \country{China}}

\author{Chong Wang}
\email{wangchong@nbu.edu.cn}
\affiliation{%
  \institution{Ningbo University}
  \city{Ningbo}
  \country{China}}

\author{Aditi Majumder}
\email{majumder@ics.uci.edu}
\affiliation{%
  \institution{University of California, Irvine}
  \country{United States}}

%


\begin{abstract}
Establishing correspondence between projector and camera images in a procam (projector + camera) system is essential for achieving high-resolution pixel matching, referred to as procam registration. The highest accuracy is typically obtained using structured light patterns (e.g., stripes or blobs). However, these methods are often inefficient and lack meaningful information for human viewers. Although some have explored the use of natural images, these often fail to provide a sufficient distribution of features to achieve comparable accuracy. Additionally, existing methods struggle to cope with environmental factors such as surface textures and variations in brightness due to ambient light or changes in camera exposure. 

To address these limitations, we propose a method based on deep neural networks. Our approach aims to generate a single natural image from text-based prompts that not only appears realistic but also possesses rich spatial features to enhance registration accuracy in procam applications. We have developed a deep neural network trained on a synthesized dataset that simulates potential geometric and photometric distortions encountered in a procam system illuminating a relatively smooth object (see Figure \ref{fig:teaser}). 

Our trained network predicts the correspondence between projector and camera images, significantly improving registration accuracy across various procam configurations. By jointly considering the naturalness and feature richness of the projector images, our method minimizes visual disruptions in projected content without sacrificing precision. A user study confirms that our technique enhances perceived naturalness and usability compared to existing methods, validating its practical utility in real-world applications.

\end{abstract}



\begin{teaserfigure}
  \centering
  \includegraphics[width=\textwidth]{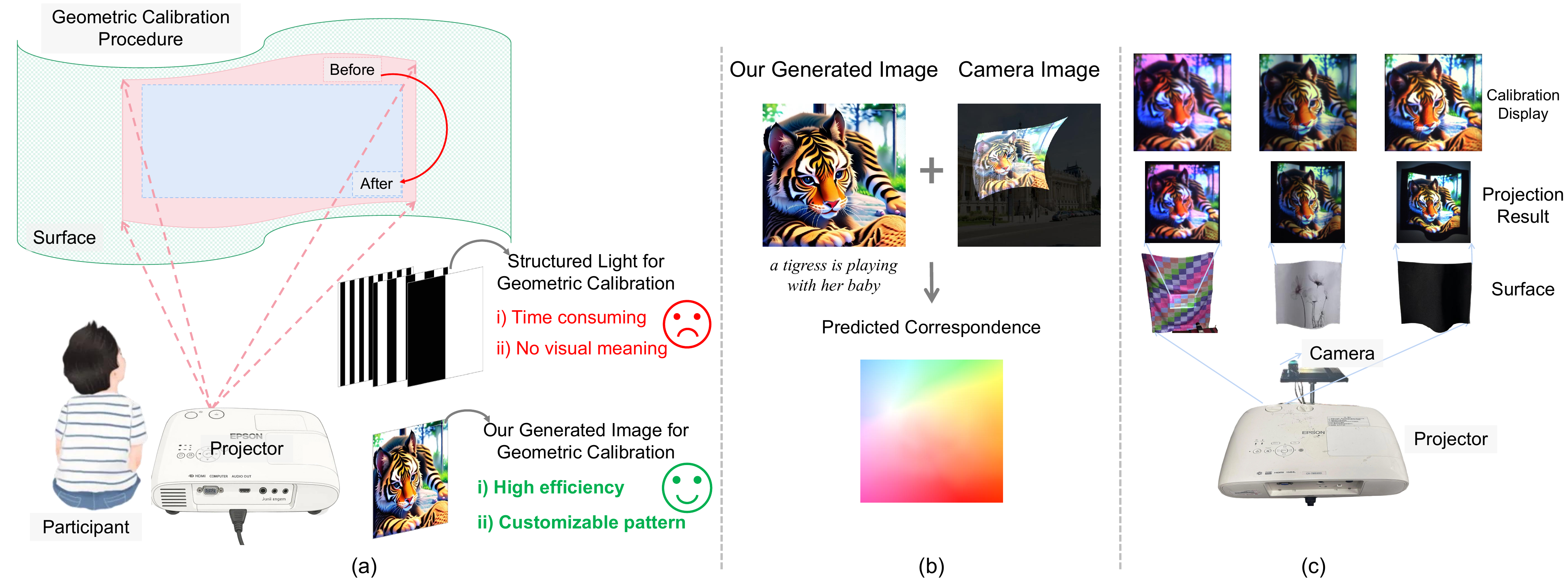}
  \caption{Text-generated non-pattern projected images are used for geometric calibration of the projector-camera system while providing a natural user experience. (a)Comparison of structured light patterns with our text-generated images. (b)We can predict the correspondence between projector and camera images by utilizing our deep Procams registration model. Here we use the color visualization method of optical flow to visualize the predicted correspondence. (c)Geometric correction for projection displays via our Procams registration model using the generated projector image.}
  \label{fig:teaser}
\end{teaserfigure}








\maketitle

\section{Introduction}

Projector-camera (Procam) systems merges the functionalities of display and capture to enable a diverse range of visualization and computer vision applications, such as spatial augmented reality (SAR)~\cite{punpongsanon2015softar,ibrahim2023spatially}, depth estimation~\cite{ming2021deep}, and spectral reflectance reconstruction~\cite{li2013robust}.  

At the core of these applications is Procam registration, which precisely aligns the projector output with the camera input. Conventional methods rely on structured light patterns (e.g., gray-code stripes \cite{cheng2024adaptive}, binary-code dots \cite{wang2020three}, phase-shifted sinusoidal code \cite{zheng2022pre}) to establish projector-camera correspondence. However, precise structured light typically requires the sequential projection of multiple patterns, leading to long acquisition times. While single-pattern structured light can alleviate time complexity, these patterns are primarily designed for machine interpretation rather than human understanding, often failing to convey visually meaningful information. This lack of aesthetic continuity can disrupt displays where visual coherence is essential. For example, although projecting structured light patterns enables precise geometric recalibration to compensate for unexpected projector displacements during movie playback, these structured patterns inevitably disrupt audience attention and compromise visual continuity‌.

Recent researchers \cite{li2023physics, debevec2023modeling, nakao2021image, oyamada2007focal} have begun seeking a solution that projects natural images and applies deep neural networks to establish the image pixel correspondence. However, the natural images currently in use are often randomly collected and may not have a nice spatial distribution of rich features  which makes them unsuitable for procam registration, resulting in a lower accuracy system. 

\subsection{Our Contribution}
We envision a system where the user can input a text that describes the ongoing display theme in text (e.g. dogs playing the woods, a majestic tiger), and the system generates an appropriate single image close to what is being displayed, but with adequate number of rich features distributed reasonably uniformly across the image. It then projects this frame consequently capturing it from the camera,  and generate the correspondences for procam registration, which can be used to warp the projector image appropriately. The controlled generation of the image provides adequate features to achieve high accuracy registration. The system is very robust to the disruption (in terms of ambient light or adjusting of camera exposures) and can enhance the user experience significantly.

To achieve the above mentioned system we propose two novel deep neural network based methods. 
\begin{enumerate}
\item 
We present a text-to-image generation method designed to generate a single natural image for Procams registration. The visual semantics of the generated images are given by users through text prompts, as shown in Fig. \ref{fig:teaser}(a). We demonstrate the first technique that leverages the distribution of local feature descriptors as a control condition to guide the generation
process from text prompts to images. The images generated by the technique are rich in spatial features, which are crucial to yield a large number of accurate correspondences enhancing their suitability for the procam registration task. We also explore the suitable distribution maps of local feature descriptors through qualitative analysis.
\item 
In order to enhance the precision of procam registration, we develop a deep neural network to predict correspondences from a pair of the projected image which is then captured by the camera. By constructing a simulation based training dataset that models the geometric and photometric distortions for camera capture of a projected image, the trained network can handle diverse environmental conditions and different procam configurations (in terms of pose, orientation, lens parameters) effectively. 
\item 
In the process, We build and release a synthesized dataset consisting of 4,500 pairs of projector and camera images under various setup configurations, providing a general-purpose benchmark for procam registration model's training and evaluation. We train a deep neural network on this dataset, effectively addressing projection distortions and camera exposures variations in both simulations and real experiments.
\item 
Experimental results demonstrate the effectiveness of our method in simultaneously expressing visual semantics and achieving precise image registration. We have also successfully applied our method to find the appropriate geometric warp for the projected image to account for the 3D surface and reconstruction of the 3D surface.
\end{enumerate}

Our method achieves a breakthrough in both accuracy and experience for using natural images in projector-camera system registration. It helps applications such as advertising design and interactive AR exhibitions get rid of traditional, visually meaningless structured light patterns, enabling more flexible and autonomous content presentation while completing geometric registration.

\section{Related Work}
\label{sec:relwork}

\subsection{Procams Registration}
Finding correspondences between the projector and the camera are crucial for (a) 3D reconstruction of the shape of the 3D projection surface; and (b) warp the projected image correctly to account for the shape of the projection surface. Over the past decades, several approaches has been designed to establish correspondences between projector and camera images. We classify them into two categories: structured-light-patterns and natural-image-patterns.

\vspace{0.5em}

{\em Structured-Light-Patterns:} Various structured light techniques have been developed focusing on different aspects like speed, accuracy, and flexibility. Structured light patterns, by definition, cannot have visual meaning to users.

Binary coded light striping~\cite{wang2020spatial} utilizes the binary codes to distinguish different light stripes, allowing easy segmentation and pattern identification. However, this method requires the projection of a large number of patterns. Gray color-coded light striping~\cite{meng2022research, al2021laser, lu2024sge} improves robustness by using Gray codes, where adjacent stripes differ by only one bit thereby reducing decoding errors. Time-coded light patterns~\cite{ye2021time} assign unique codes to each stripe over time, requiring multiple projections to decode spatial information, while spatial codification projects a unique spatial pattern to handle moving objects with more complex decoding.

Phase shift methods~\cite{zheng2022pre, kim2023design, zeng2022self, pak2021verification} project sinusoidal patterns to interpolate between adjacent light planes, offering high-resolution results with sub-pixel accuracy, but are limited to static scenes. De Bruijn sequences~\cite{marcovich2021balanced} and M-arrays~\cite{zhou2021m} provide methods for handling occlusions by ensuring each pattern region is unique, but involve complex decoding. Techniques like direct encoding with color use color variations for pattern encoding~\cite{kayhan2021content, abdou2021can}, allowing for easy matching under the assumption that the scene does not alter the projected colors (e.g. via ambient lighting). These techniques each come with specific advantages and drawbacks, such as high resolution and accuracy but inapplicability to moving objects for time-multiplexing, or suitability for moving objects but lower resolution and complex decoding for spatial codification.

Some other works focus on imperceptible patterns for geometric correction and projection display simultaneously. These methods involve integrating hidden infrared (IR) lights~\cite{hashimoto2017dynamic} or synchronization triggers~\cite{10.5555/1770090.1770106} to ensure that the structural patterns remain undetectable to human observers while being captured by cameras. However, the additions of these extra hardware raises both implementation expenses and operational challenges. Recently, adaptive patterns~\cite{dong2023adaptive} have been designed to improve the precision of procam registration progressively. The adaptive patterns are generated iteratively to maximize color discrimination for decoding in the presence of ambient light. 

\vspace{0.5em}

{\em Natural-image-Patterns:} The earliest natural-image-based approach~\cite{takahashi2008geometric} emerges to address the challenges of continuous calibration on moving planar surfaces. This method extracts local features, such as SIFT and SURF, from the display's visual content, and estimates the homography matrix between the projected and camera-captured images to correct the display. Next, we see advent of similar methods for non-planar surfaces that utilize optical flow algorithms, such as Lucas-Kanade~\cite{zollmann2007passive}, to obtain the pixel correspondences. Other methods use the assumption of a static procams system and utilize the epipolar constraints to predict the offset of the inter-pixel correspondence in real time~\cite{hashimoto2021radiometric}. However, these numerical optimization-based correspondence estimation methods approaches cannot achieve high accuracy since they struggle to handle the nonlinear responses of devices and textured surfaces.

Deep neural networks' power to handle such complex systems makes them appropriate to handle these issues. Huang \textit{et al.}~\cite{huang2019compennet++} design a warping network that consists of a coarse transformation using affine mapping and a refinement transformation using thin plate spline (TPS) mapping. This method optimizes the trainable parameters of both transformations by feeding batches of image pairs of projector and camera into the network, to obtain the pixel correspondences.  However, procams warp is formulated as a high-dimensional fitting problem in this method, requiring a significant amount of time to converge. Li \textit{et al.}~\cite{li2023physics} also use natural-images-based warp to decouple the factors of surface textures and geometric distortion via alternative optimization. It utilizes a deep optical flow network named GMA to predict the displacement between the pixels of projector images and camera images in the presence of the surface textures. The primary strength of this approach lies in its real-time efficiency and robustness in contrast to prior methods.  Kageyama \textit{et al.}~\cite{kageyama2024efficient} retrains the GMA optical flow network and a PSF estimation network on a synthesized dataset to handle both out-of-focus blur and spatial distortion. However, these cannot guarantee high accuracy due to lack of rich spatial features in the natural image. 

\subsection{Text-to-Image Generation}
Text-to-Image generation techniques create images with desired visual semantics from given text prompts. The most successful image generative models are diffusion-based~\cite{yang2023diffusion}. These models frame the generation process as the reverse of a diffusion process, iteratively generating images from noise. To reduce the dimension of the data and enhance the control flexibility of the generation, latent spaces are introduced to diffusion models~\cite{rombach2022high}. This approach is effective in handling the complexity and diversity of different data modalities~\cite{ma2023unified}. To further refine control over the image generation process according to user-specified inputs or conditions, such as edge maps and depth maps, a ControlNet module is proposed for integration into the base diffusion model, the main motivation for our proposed method. While some works~\cite{liao2024diffusion} focus on generating non-pattern QR-codes without compromising the ability to scan them, previous methods have not explored creating natural images with rich spatial features for procam systems.

\section{Text-to-Image Generation}
\label{sec:textimage}
We hypothesize that the quality of procam registration is dependent on the richness (like number, distribution, ability to standout)  of the spatial features in the projector image. Therefore, we aim to generate a natural image according to the given text prompt description ensuring the image has abundant spatial features.

\subsection{Feature Distribution}
The most straightforward approach to adding conditional control to image generation is to ask users to provide additional information that directly specifies the desired output image. This allows the algorithm to generate images based on the user's specified conditions or requirements. We propose to add a feature distribution map to control the generation. The used feature can be either the hand-crafted feature or the learned feature. For instance, we use a feature map in the spatial attention block of our deep registration network (Section \ref{sec:procamreg}) effectively captures the distribution.  In this paper, we apply ORB (Oriented FAST and Rotated BRIEF) feature descriptor~\cite{rublee2011orb} to generate the feature distribution of projector images. 

Consider a projector image denoted as $\mathbf{I}_p$. Let us extract its ORB features $\{(\mathbf{c}_i, \sigma_i)|i=1,\dots,k\}$,where $k$ represents the number of the extracted features. Here, $\mathbf{c}_i$ and $\sigma_i$ denote the coordinate and scale of the $i$-th feature respectively. The feature distribution map $\mathbf{M}$ is created by:
\begin{equation}
    \mathbf{M}_p(\mathbf{x}) = \sum_{i=1}^{k} exp(-\frac{||\mathbf{x}-\mathbf{c}_i||_2^2}{\sigma_i^2}),
\end{equation}
where $\mathbf{x}$ denotes the coordinate of a pixel in the feature distribution map $\mathbf{M}$. Here we treat each ORB feature point as a Gaussian blob to obtain the feature distribution map. The resolution of the feature distribution map is the same as the projector image. The primary reason for using ORB is its demonstrated effectiveness and efficiency in previous Procams registration work~\cite{li2012real}. Furthermore, the ORB descriptor's independence from deep registration networks enhances its adaptability to a wide range of registration techniques. We show examples of the feature distribution map in Fig. \ref{fig:controlnet}.

\begin{figure}[h]
 \centering 
 \includegraphics[width=0.8\columnwidth]{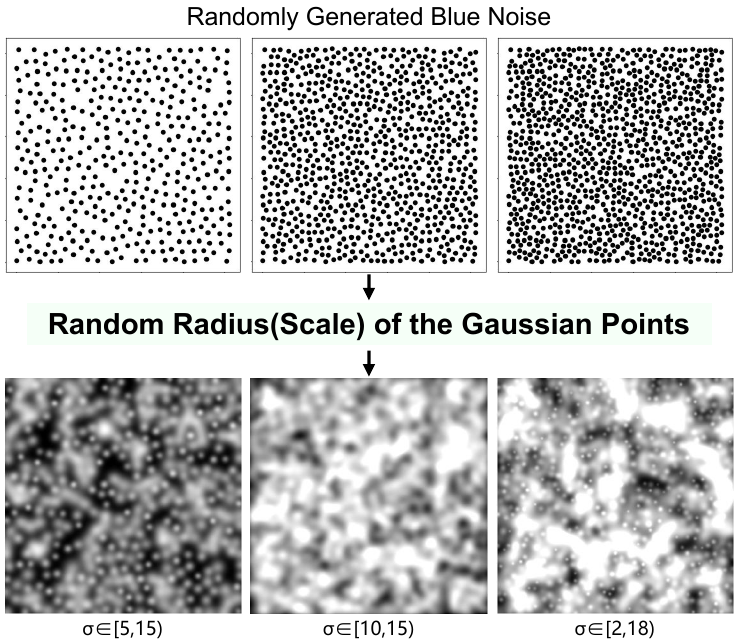}
 \caption{The generation process of Gaussian distribution map firstly generates different number of blue noise maps randomly, and then randomly assigns the radius of Gaussian sphere on the basis of these blue noise maps, and finally generates the randomized Gaussian distribution map ($\sigma$ represents the scale of the Gaussian points).}
 \label{fig:orb}
\end{figure}

By employing the feature distribution map as conditional control data, we develope a text-to-image generation network tailored for natural image creation. This enables us to produce images with desired spatial feature distributions by providing corresponding feature distribution maps.

\subsection{Image Generation}
\begin{figure*}[h]
 \centering 
 \includegraphics[width=\textwidth]{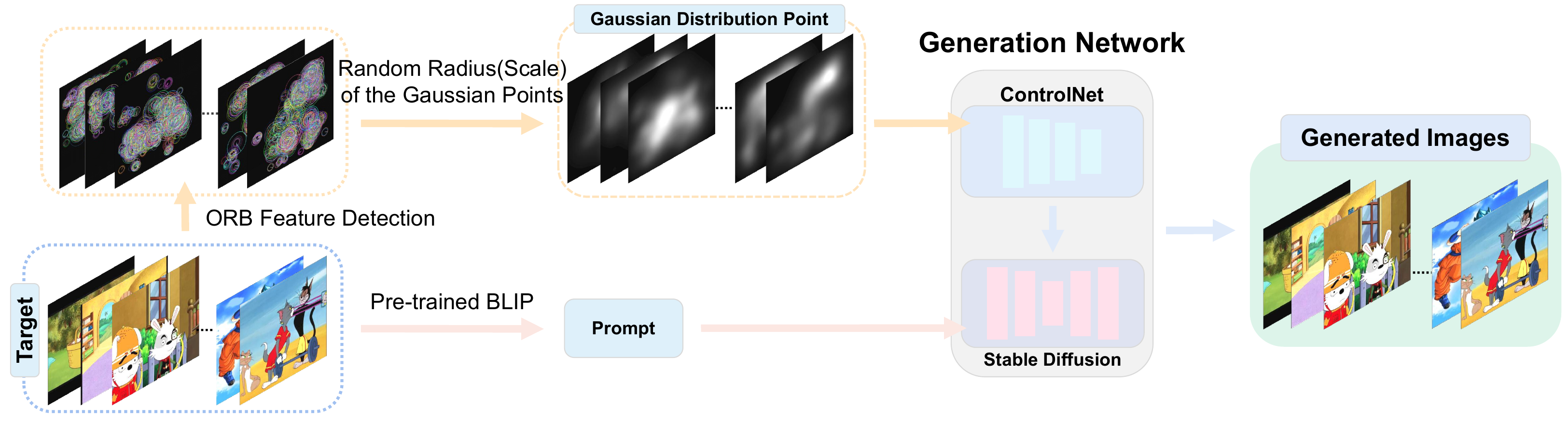}
 \caption{The control network architecture for natural image generation. The network consists of two parallel branches: one extracts image features using ORB corner detection and converts them into a Gaussian distribution, while the other extracts image description information through a pre-trained BLIP model. The feature distribution and text prompts are then fed into ControlNet and Stable Diffusion for training. During training, the parameters of the pre-trained model are fixed, and only the control module is optimized to maintain generation quality and prevent knowledge degradation.}
 \label{fig:controlnet}
\end{figure*}

The control network we used for natural image creation is ControlNet~\cite{zhang2023adding}. We fed the extracted spatial features into ControlNet as conditional inputs. These features serve as a guide, instructing the model on how to generate images that adhere to this distribution. We select Stable Diffusion~\cite{rombach2022high} as our generation network. The used network architecture for conditional control and image generation is shown in Fig. \ref{fig:controlnet}. Rather than training from scratch, we utilize a large pretrained model~\cite{LibLib2023} that serves as a robust backbone for image generation, having been trained on extensive datasets and possessing diverse capabilities. During training, we lock the parameters of the pretrained model and progressively optimize the control module to to preserve generation quality and prevent the knowledge it contains from potential degradation caused by the control conditions. 

In the image generation network, given an initial input image \( z_0 \), the algorithm iteratively adds noise to generate images \( z_t \) with varying noise levels, where \( t \) represents the number of noise addition steps. The algorithm operates under conditions such as the time step \( t \), a text prompt \( c_t \), and an encoded vector \( c_f \) of the task-specific conditions \( \mathbf{M} \). Its objective is to train a network with parameters \( \lambda_\theta \) to predict the amount of noise added to the noisy image \( z_t \):
\begin{equation}
    \mathcal{L}=\mathbf{E}_{x_{0}, t, c_{t}, c_{f}, \epsilon \sim \mathcal{N}(0,1)}\left[\left\|\epsilon_{\theta}\left(x_{t}, t, c_{t}, c_{f}\right)-\epsilon\right\|_{2}^{2}\right].
\end{equation}
where $\mathcal{L}$ is the overall learning objective of the entire image generation network.

Once the generation network with feature distribution control is trained, we provide prompts and distribution maps to generate images. Ideally, the features in the distribution map should exhibit high density and uniformity. To accomplish this, a blue noise sampling strategy is employed to determine the spatial positions of the features, ensuring a feature map with a uniform distribution. Initially, dense points are randomly sampled, and points are iteratively eliminated to ensure a sufficiently large minimum distance between them. We present examples of feature distribution maps in Fig. \ref{fig:orb}. 

\begin{figure*}[tb]
 \centering 
 \includegraphics[width=\textwidth]{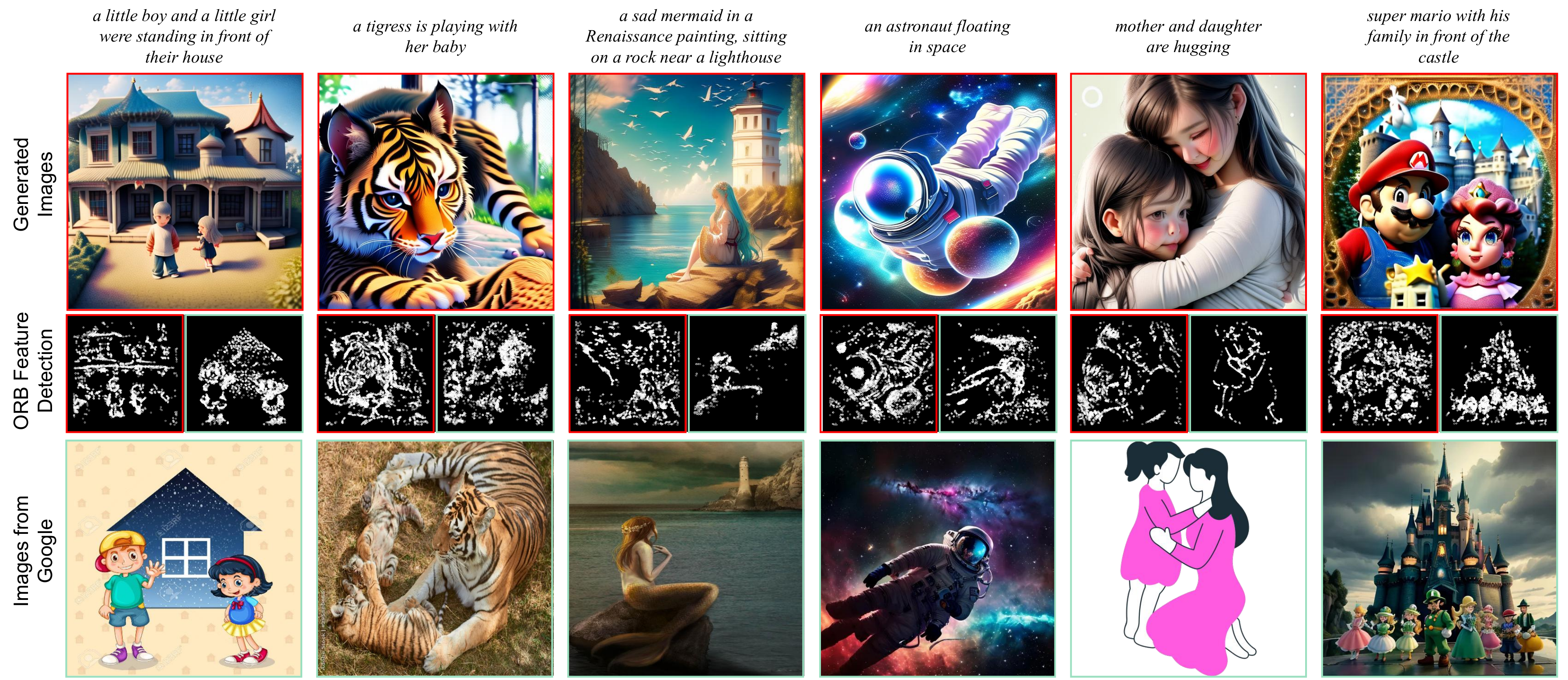}
 \caption{The figure displays images generated using our proposed non-pattern projector image generation technique, alongside images downloaded from Google using the same text prompts. \textbf{Note:} The red box represents the generated projected image, the green box indicates the image obtained from Google, and the center displays the feature points extracted using ORB feature detection.}
 \label{fig:showimages}
\end{figure*}

We generated several images with a resolution of $512 \times 512$ using randomly generated Gaussian feature distributions and textual prompts, some of which were generated as shown in Fig. \ref{fig:showimages}. The images are generated using our method with the feature sampling density of $0.02$ and the scale range of $[31.00, 111.08]$.

Fig. \ref{fig:showimages} shows the results of our image generation technique and compares it with the first image returned by google search using the same text prompt. For each textual prompt, we searched Google using the same keywords and downloaded the first $512 \times 512$ image it returned as comparison images. It is evident that the orb features of the generated images are distributed more uniformly than those of the Google images. We will investigate the influences of the intensity and density of the spatial features in Section \ref{sec:sim}.

\section{Deep Procams Registration}
\label{sec:procamreg}
Previous works use pre-trained deep supervised optical flow estimation neural networks to predict pixel displacements in procam image pairs. However, the dataset for optical flow estimation contains only temporally consecutive images with small motions under similar lighting conditions. On the contrary, in an arbitrarily deployed procam system, the pixel displacement between the camera image and the projector image shows much larger geometric distortion and photometric change. Additionally, the camera-captured image is influenced not only by the content of the projected image but also by the texture of the projection surface. This is usually due to ambient light that causes brightness variation which in turn can cause the camera to apply an automatic exposure change in the captured image. These factors are not taken into account in previous optical flow networks, like GMA. 

Acquiring real image pairs and the ground truth of the registration for learning is a huge amount of work. Therefore, we introduce a synthesized dataset and then train a deep supervised neural network on the dataset to obtain the model for procam registration.

\subsection{Dataset}
\begin{figure*}[h]
 \centering 
 \includegraphics[width=\textwidth]{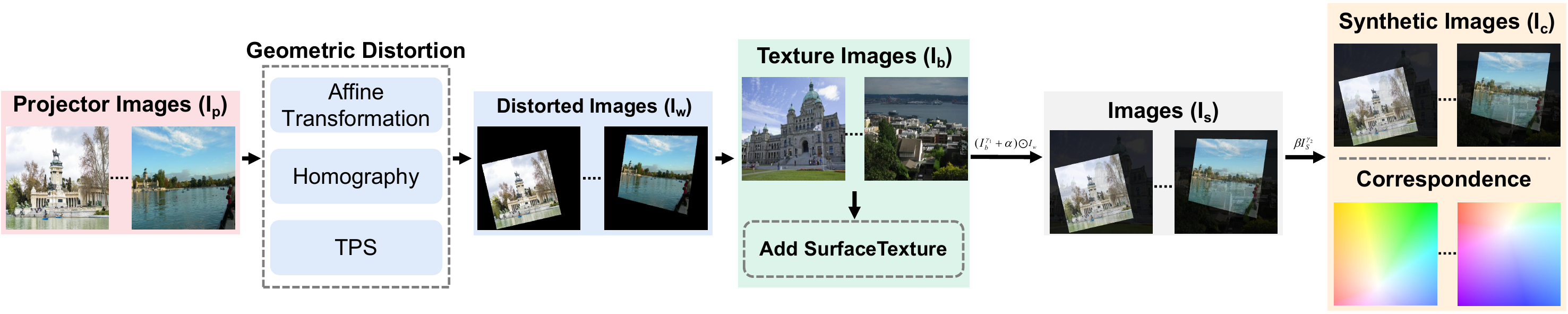}
 \caption{The figure illustrates the overall framework of the data synthesis process. On the left, randomly selected natural images (that can be of scenes or can include projected images) undergo geometric transformations such as affine, homography, and thin-plate spline (TPS) transformations. The texture images are then added to the transformed images to enhance realism of the final synthesized images. On the right, the synthesized images (captured images) and their corresponding flow ground truth are shown, which will be used for model training and evaluation.  Here we use the color visualization method of optical flow to visualize the predicted correspondence.}
 \label{fig:Geotransform}
\end{figure*}

The imaging process of procams is significantly affected by various factors, including surface geometry and reflectance~\cite{li2023physics}. To simulate the images obtained during actual imaging, we reproduce distortions, exposure fluctuations, textures, and other factors in the camera images within the synthesized dataset. Without loss of generality, let's assume a projector image denoted as $\{\mathbf{I}_p\}$ and the corresponding synthesized camera image denoted as $\{\mathbf{I}_c\}$. The synthesis process is shown in Fig. \ref{fig:Geotransform}.

\vspace{0.5em}

{\em Simulating Distortion:} The distortion observed in the procam image pairs typically arise from factors such as the non-coaxial structure of procam setup, surface geometry, and lens aberrations. Previous studies~\cite{kageyama2024efficient, huang2019compennet++} have suggested that this distortion can be effectively modeled by a combination of homography mapping and nonlinear Thin Plate Spline (TPS) transformation. Building upon this foundation, we have employed three types of transformations to distort the projector image $\mathbf{I}_p$: affine, homography, and TPS transformations. While using solely affine and homography transformations is suitable for planar surfaces, integrating TPS transformation enables handling smooth non-planar surfaces as well. We randomly sample parameters for three types of transformations, synthesize distorted images of $\mathbf{I}_p$, and then select distorted images that the content does not exceed the valid range and the image area does not significantly change to form the set $\{\mathbf{I}_w\}$.

\vspace{0.5em}

{\em Simulating Surface Texture:}  To simulate the impact of surface reflectance on camera images, we introduce another image $\mathbf{I}_b$ as a reflectance map. Note that $\mathbf{I}_b$ has three channels(R, G, B), where the intensity values of each pixel in every channel represent the reflectance ratios of that pixel in the respective channel. We also consider ambient light since it plays a crucial role in determining the quality of the projected image as well as the performance of the camera capturing the scene. Therefore, we add the effects of surface texture and ambient light to the warped images to get $\mathbf{I}_s$:
\begin{equation}
   \mathbf{I}_s = (\mathbf{I}_b^{\gamma_1} +\mathbf{\alpha}) \odot \mathbf{I}_w,
\end{equation}
where $\alpha$ is a parameter to adjust the intensity of the ambient light, and $\gamma_1$ denotes the exponential of the gamma-shaped response function of the projector. Here image $\mathbf{I}_s$ is the corresponding synthesized reflective irradiance of the projector image $\mathbf{I}_p$.

\vspace{0.5em}

{\em Simulating Exposure Variation:} As the exposure settings and response functions of the camera vary, we utilize a parameter $\beta$  to adjust the exposure of the captured image within the synthesized dataset. Furthermore, to simulate cameras with different response functions, we employ a parameter $\gamma_2$, representing the exponential of the gamma-shaped response function of the cameras. Consequently, the formulation for the synthesized camera images $\mathbf{I}_c$  is as follows:
\begin{equation}
   \mathbf{I}_c = \beta \, \mathbf{I}_s^{\gamma_2}.
\end{equation}

We collect the image pairs of  $\mathbf{I}_b$ and $\mathbf{I}_c$ to construct the synthesized dataset for the training of deep registration network. The parameter $\gamma_1$ is randomly set to the range of $[1.5, 2.5]$, $\gamma_2$ is randomly set to the range of $[0.4, 0.66]$ and the exposure parameter $\beta$ is randomly set to the range of $[0.8, 1.25]$.

\subsection{Deep Registration Network}

The objective of the deep registration network is to establish robust correspondences between projector-captured images and their associated camera-captured images. Conventional CNN-based optical flow estimation approaches frequently encounter challenges in handling scenarios characterized by substantial differences and large displacements between the image pairs. Therefore, a transformer-based network that applies a global self-attention mechanism to extract features is more suitable in this task. In addition, due to the difference between the image pairs, inter-attention across the two images is essential for extracting and aligning the feature maps.

\begin{figure}[h]
 \centering 
 \includegraphics[width=\columnwidth]{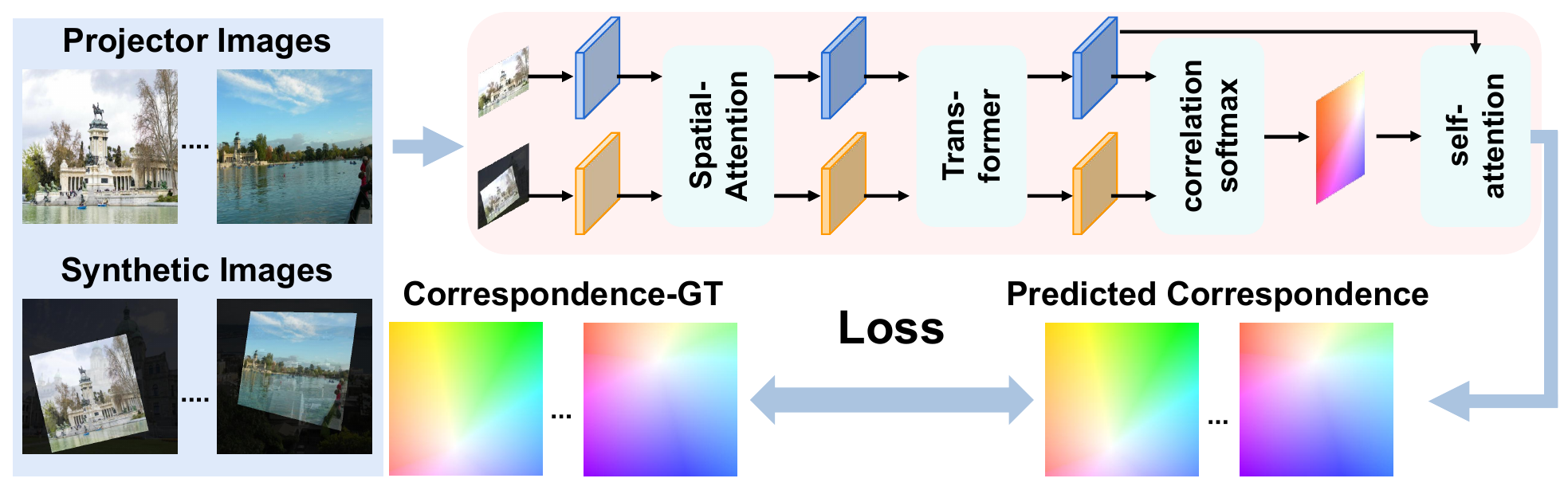}
 \caption{Framework of the Deep Registration Network: The network extracts deep features from the projected and synthetic images using weight-shared convolutional blocks. On this basis, the Transformer module and Spatial-Attention mechanism further enhance the similarity between feature mappings and focus on the spatial features within the images, thereby achieving precise registration.}
 \label{fig:flownet}
\end{figure}

\begin{figure*}[tb]
 \centering 
 \includegraphics[width=\textwidth]{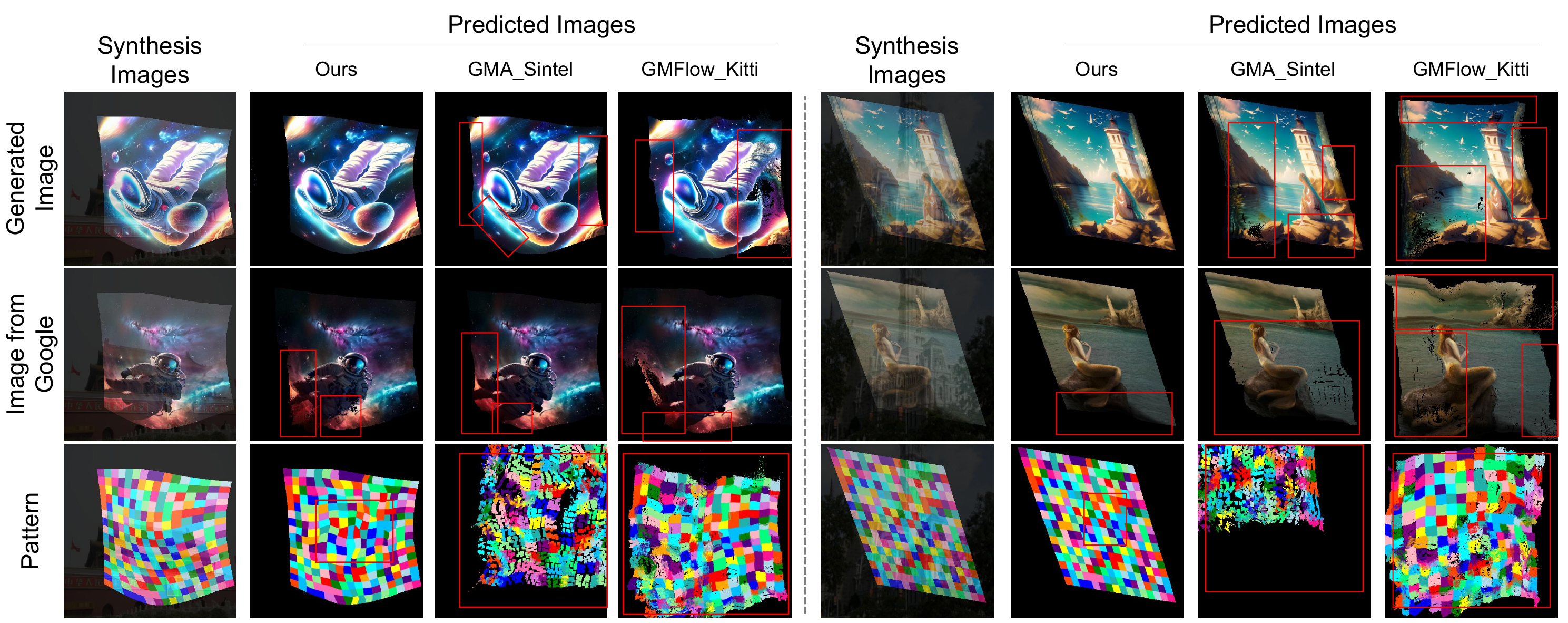}
 \caption{The warped results obtained by applying the predicted correspondence from different models to generated projector images, Google images, and structured pattern. \textbf{Note:} The red box highlights areas where the correspondence results are significantly not correct.}
 \label{fig:simulation}
\end{figure*}

Based on these considerations, a transformer-based network inspired by GMFlow framework\cite{xu2022gmflow} is proposed for learning the image registration process. As shown in Fig. \ref{fig:flownet}, we first extract two deep features from the image pairs using a weight-sharing convolutional block. In this block, we utilize the spatial attention mechanism to focus more on the spatial features within the block. The attention map generated through this mechanism helps to explore the richness of spatial features for the purpose of registration. However, the two feature maps may not be similar due to the effects of background textures and exposure variation. To address this, we utilize a transformer~\cite{liu2021swin} to refine the extracted feature maps. This allows for better feature matching by considering the cross-channel attention between the two feature maps. In the correspondence estimation process, we use an additional self-attention layer to propagate accurate predictions from matched pixels to unmatched ones, taking into consideration the feature self-similarity. Finally, we train the deep network by minimizing the L2-norm loss function between predicted correspondences and ground truth to accomplish procam registration.

\begin{table*}[h]
\centering
\caption{The evaluation results of the end-point error (error) for the six best-performing models out of the eight, alongside our model.  This table presents pixel alignment results for each model and compares alignment performance using non-pattern projector images and google images as projection images.  The maximum error represents the highest error observed across 14 test sets, while the mean and standard deviation are calculated from the error results of these test sets.}
\label{tab:compare_with_other_model}
\renewcommand{\arraystretch}{1.2} 
\resizebox{\textwidth}{!}{%
\begin{tabular}{lcccccccccccccccccccccc}
\toprule
\multirow{2}{*}{} & \multicolumn{3}{c}{\textbf{GMA\_Chairs}} &  \multicolumn{3}{c}{\textbf{GMA\_Sintel}} & \multicolumn{3}{c}{\textbf{GMA\_Things}} & \multicolumn{3}{c}{\textbf{GMFlow\_Kitti}} & \multicolumn{3}{c}{\textbf{GMFlow\_Sintel}} & \multicolumn{3}{c}{\textbf{GMFlow\_Things}} & \multicolumn{3}{c}{\textbf{\emph{Ours}}}\\
\cmidrule(r){2-4} \cmidrule(r){5-7} \cmidrule(r){8-10} \cmidrule(r){11-13} \cmidrule(r){14-16} \cmidrule(r){17-19} \cmidrule(r){20-22}
& MAX & MEAN & SD & MAX & MEAN & SD & MAX & MEAN & SD & MAX & MEAN & SD & MAX & MEAN & SD & MAX & MEAN & SD & MAX & MEAN & SD \\ 
\midrule
 Generated Images & 73.809 & 53.922 & 11.401 & 15.884 & 7.530 & 4.504 & 42.236 & 17.506 & 10.531 & 20.390 & 16.633 & 3.310 & 25.095 & 12.630 & 4.720 & 30.847 & 18.212 & 6.756 & \textbf{3.584} & \textbf{2.956}& \textbf{0.386} & \\
Images from Google & 92.248 & 79.982 & 8.428 & 67.033 & 34.376 & 17.033 & 91.961 & 55.750 & 17.925 & 48.404 & 31.832 & 10.814 & 97.773 & 38.575 & 24.350 & 119.537 & 62.558 & 31.851 & \textbf{33.616} & \textbf{6.287} & \textbf{8.016} & \\
\cmidrule(r){2-4} \cmidrule(r){5-7} \cmidrule(r){8-10} \cmidrule(r){11-13} \cmidrule(r){14-16} \cmidrule(r){17-19} \cmidrule(r){20-22}
Pattern &  & 108.986 &   &  & 220.241 &  &  & 187.870 &  &  & 35.043 &  &  & 74.880 &  &  & 79.796 &  &  & \textbf{2.092} &  & \\
\bottomrule
\end{tabular}%
}
\end{table*}

\section{Results in Simulation}
\label{sec:sim}
Armed with the text to image generation and the procam registration dataset, we now run test in simulation and using a real procam setup. In this section we detail these results.

We constructed image datasets for testing and conducted simulation experiments to evaluate the effectiveness of our image registration network with our text prompted projector images. We compared the pixel alignment accuracy between images generated by our models (Section \ref{sec:textimage}) and those sourced from Google. Additionally, we used a pattern image(see Fig. \ref{fig:simulation}) suggested by \cite{erel2023neural} as another comparison image. 

To compare the performance of our generated single images, general images, and color pattern on procam registration, we synthesize the test dataset of image registration by applying the geometric transformation, texture, and exposure variation to the three types of images as previously mentioned. The texture images we used are selected from the MegaDepth\_CAPS \cite{li2018megadepth} dataset. Note that the synthesis method for training set and testing set are the same while with different parameter configurations such as spatial transformation and exposure values. The testing set consists of the synthesized images with 150 sets of parameter configurations and their corresponding displacement. We evaluate our generated images and deep procams registration network on the testing dataset.

We use the pixel offset accuracy as the metric of alignment for comparison. We also systematically analyze the effect of the quantity and scale of feature points on projector image alignment  by comparing our registration model with others. The error metric we used is the L2 norm of the difference between the predicted correspondence displacement and ground truth ones.  

{\em Implementation Details:} For the GMA\cite{jiang2021learning} and GMFlow~\cite{xu2022gmflow} models, we used the same code and experimental settings as the official implementations. To ensure fairness, factors such as exposure parameters and the position of projected images were kept consistent across experiments. All training was conducted on an NVIDIA RTX 4090 GPU. The generative network was trained for 49 epochs, while the deep registration network was trained for 227 epochs.

Table \ref{tab:compare_with_other_model} presents the comparisons of the error among the three types of images using our registration prediction models and eight others. Among them, four models used are pre-trained GMFlow models \cite{xu2022gmflow}, and the other four are pre-trained GMA models proposed by Jiang et al. \cite{jiang2021learning}. This demonstrates convincingly that our models higher pixel alignment accuracy due to more detectable feature points in the text prompted image generation and exhibit less error across all models. Compared to Google images, our generated projector images achieve higher accuracy in pixel alignment with lower error, further demonstrating their suitability for precise registration tasks.
This is true even when the procam registration is applied to images from diverse sources, whether they are generated images, google images, or color patterns. Furthermore, the use of our generated projector images consistently results in the lowest pixel error across other models as well.

We visualized the results of the two models with the lowest error (GMA\_Sintel and GMFlow\_Kitti) alongside the Procams registration model, as shown in Fig. \ref{fig:simulation}. The results indicate that the warped images predicted by our model are the closest to the original synthetic images (ground truth). Other registration models, such as GMA and GMFlow, fail in the given cases due to the influence of the background textures.

\begin{table}[htb]
\centering
\caption{The experimental results of the Procams registration model compared with other models in real-world scenarios are presented. The table shows the error metric results for the four best-performing models out of the eight, alongside our model.}
\label{tab:real}
\renewcommand{\arraystretch}{1.5} 
\resizebox{\columnwidth}{!}{%
\begin{tabular}{cccc}
\toprule
& \multicolumn{3}{c}{\textbf{Error}}   \\
\cmidrule(r){2-4}  
& Generated Images & Images from Google & Pattern\\ 
& Scene 1/Scene 2 & Scene 1/ Scene 2 & Scene 1/Scene 2 \\
\midrule
GMA\_Sintel & 21.952/22.211 & 57.376/66.430 & 298.004/250.301 \\
\midrule
GMA\_Things & 41.658/44.859 & 95.916/92.774 & 133.563/201.017 \\
\midrule 
GMFlow\_Chairs & 110.540/105.869 & 126.340/121.230 & 113.719/114.553\\
\midrule
GMFlow\_Things & 154.179/144.620 & 164.083/151.523 & 242.583/201.259\\
\midrule
Ours & \textbf{5.222/12.139} & \textbf{9.238/13.834} & \textbf{1.858/6.796}\\
\bottomrule
\end{tabular}
}
\end{table}

\begin{figure}[tb]
 \centering 
 \includegraphics[width=\columnwidth]{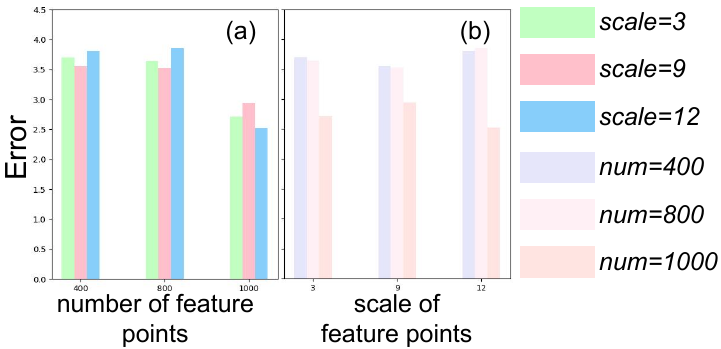}
 \caption{Visualize the errors of the sample according to the number and scale of input Gaussian feature points. (a) presents a bar chart of the error metrics obtained by varying the scales of feature points while keeping the number of feature points constant. (b) illustrates a bar chart of the error metrics obtained by varying the number of feature points while maintaining a consistent scale.}
\label{fig:box}
\end{figure}

We further investigate the performance of the generated projector images by systematically varying the quantity and scale of the features used. To conduct our experiments, we utilized a single text prompt: (``two cute rabbits''). We tested three different quantities of Gaussian feature points: 400, 800, and 1,000 points. For each quantity, we applied three different scale conditions: scale=3, scale=9, and scale=12. Given that the resulting data did not follow a normal distribution, we used Spearman's correlation analysis for qualitative assessment. As shown in the Table \ref{tab:spearman} and Fig. \ref{fig:box}, we found a significant negative correlation between the number of feature points and pixel alignment error, indicating that as the number of feature points increases, pixel alignment error tends to decrease significantly. Additionally, in this preliminary experiment, the results showed that scale did not have a significant impact on the performance of the projector images. 

\begin{table}[]
\centering
\caption{The Spearman analysis revealed the impact of varying the number and scale of Gaussian feature points on pixel alignment error. P represents the significance value and \(\tau\) represents the Kendall correlation coefficient. The symbols ***, **, and * correspond to significance levels of 0.1$\%$, 1$\%$, and 5$\%$, respectively.}
\renewcommand{\arraystretch}{1.2} 
\label{tab:spearman}
\begin{tabular}{ccccc}
\hline
                            &       \multicolumn{2}{c}{Points' Number} & \multicolumn{2}{c}{Points' Scale} \\
                            &       \emph{p}       & \emph{ \(\tau\)}      & \emph{p}        & \emph{ \(\tau\)}      \\
                            \hline
     Error & 0.042* & -0.685 & 0.258 & -0.422 \\ 
\hline
\end{tabular}
\end{table}

      


\begin{figure*}[htp]
 \centering 
 \includegraphics[width=\textwidth]{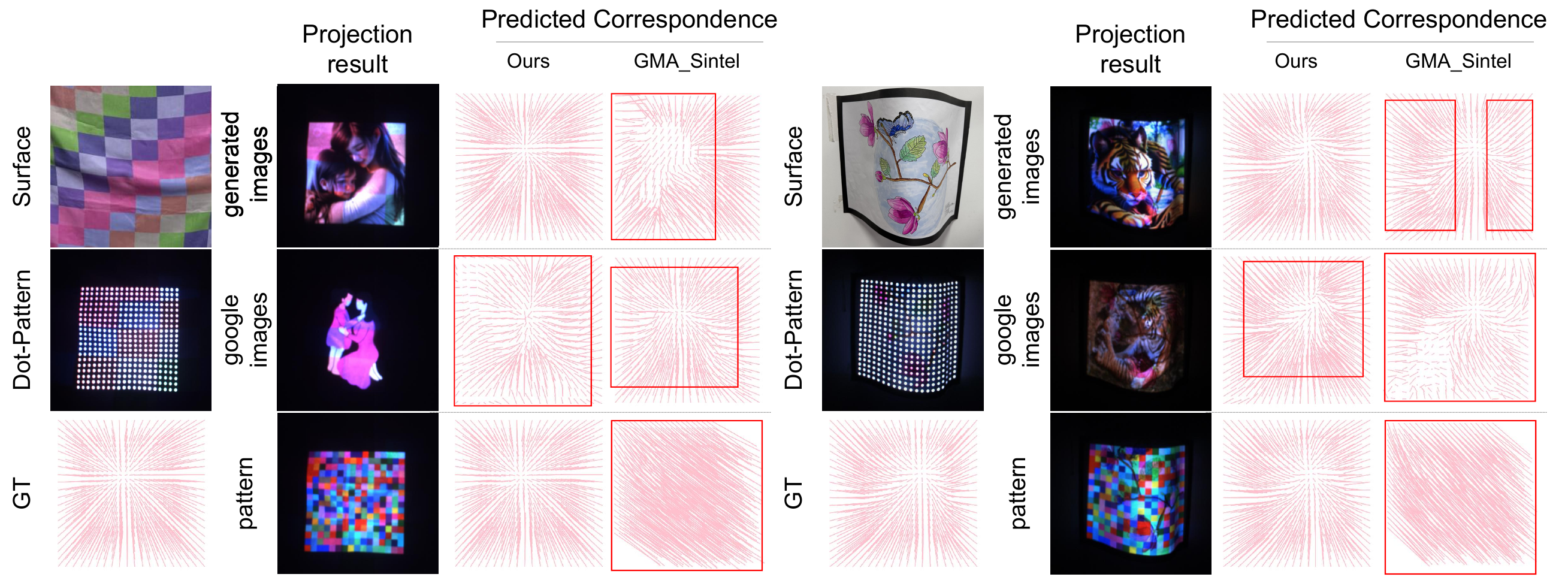}
 \caption{Projecting images onto different textured objects and comparing the predicted correspondence results of the GMA\_Sintel model with our model. To obtain the ground truth for correspondence, we projected binary dot matrix images ($20 \times 20$ dots) onto the surfaces of two scenarios. By decoding the coordinates of these points in both the projected and captured images, we derived the correspondence data to use as ground truth.  \textbf{Note:} The red box highlights areas where the correspondence results are significantly not correct.}
 \label{fig:experiment}
\end{figure*}

\section{Results in a Real Setup}
\label{sec:exp}

Based on the results of the simulation, we conducted real-world experiments to verify the robustness and effectiveness of our model. These experiments aim to evaluate the model's performance in actual projection environments, laying a solid foundation for the practical application of our procam registration technology. 

{\em Setup:} We used a high precision laser projector to project images onto the surfaces of objects without out-of-focus bluring effect and captured these images with a high resolution industrial camera ($2448 \times 2050$). To ensure that the relative positions of the camera and projector remained constant throughout the experiment, both were securely fixed in place, with no adjustments made during the entire process, as shown in Fig. \ref{fig:teaser}(c). This setup provided a high level of stability and accuracy during image acquisition. 

{\em Experiments:} In the experiment, we first projected binary dot matrix images ($20 \times 20$ dots) to serve as the ground truth. We then projected the generated images and images downloaded from Google onto the surfaces of the objects. The objects used in the experiment included a plaid(scene 1) and a art painting (scene 2), representing simple and complex geometric shape/texture, respectively.

Finally, we input these projected images and the corresponding captured images into a deep registration network to predict the correspondence and generate the corresponding pixel coordinates. We then compared these predicted pixel coordinates with the points at the same positions in the ground truth to evaluate the accuracy of our model's predictions. We compared the pixel displacement predictions of our procam registration model with the other eight models mentioned in the simulation experiments when projecting images onto real objects (see Fig. \ref{fig:experiment}).

{\em Results:} The results shown in Table \ref{tab:real} indicate that our procam registration model consistently outperformed the other models, achieving lower pixel errors on both the generated projector images and the images downloaded from Google. Notably, when using the our image generation technique, the procams registration model delivered the most accurate pixel alignment results among all models.

The experimental results reveal a trend in pixel error variation based on the geometric complexity and texture features of the object's surface. Specifically, projecting images onto simpler textures and surfaces significantly reduces pixel error compared to more complex models. This indicates that geometric complexity and texture features have a significant impact on pixel alignment accuracy, with simpler geometries and textures reducing the error.

\begin{figure*}[h]
 \centering 
 \includegraphics[width=\textwidth]{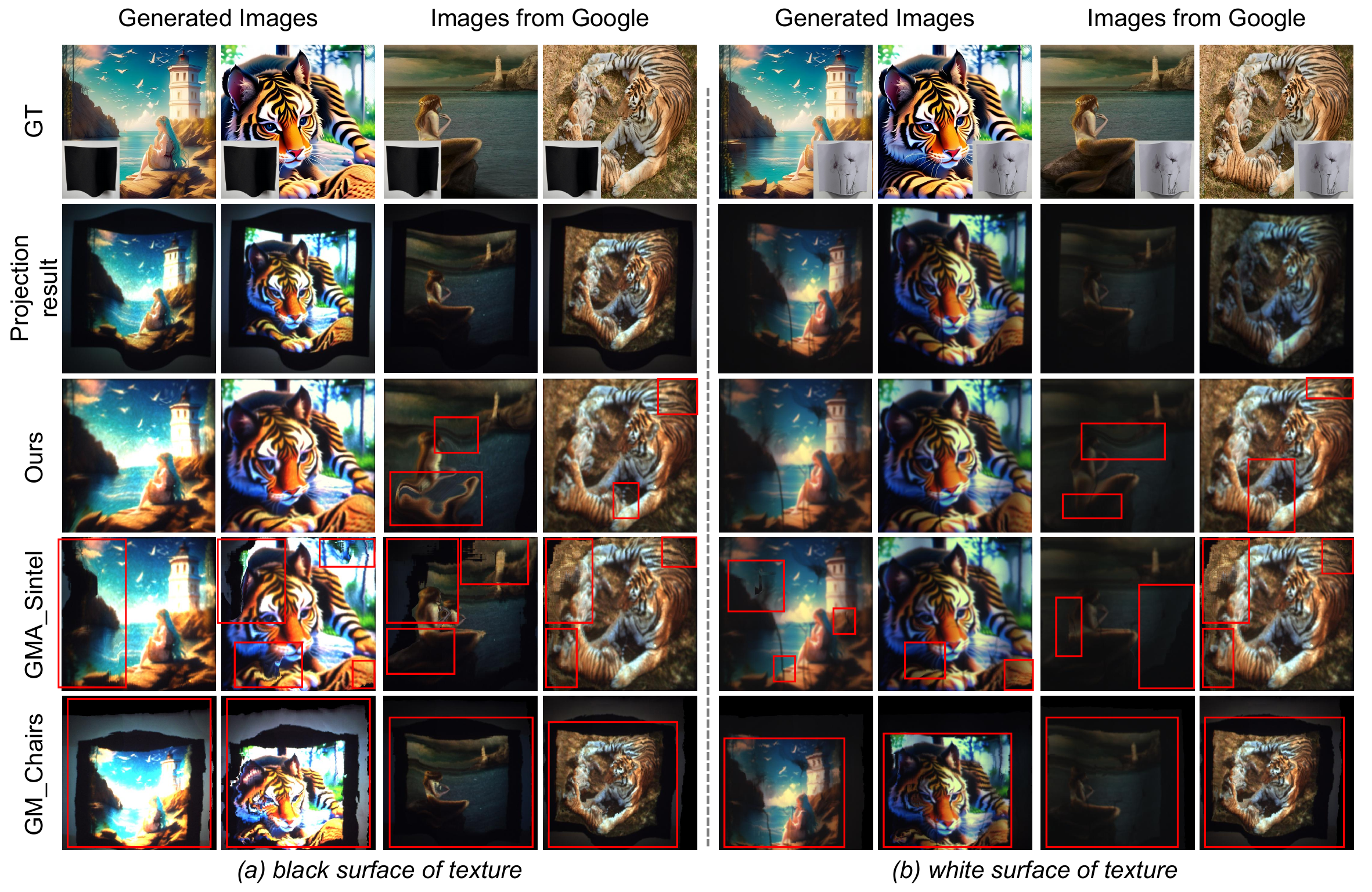}
 \vspace{-2.5em}
 \caption{Geometric calibration results after projecting images onto surfaces with different textures. We compared five models (including our method, GMA\_Sintel, GMA\_Things, GM\_Chairs, and GM\_Things) by evaluating the calibration performance on generated projected images and images downloaded from Google, assessing the results on both smooth and textured surfaces.  \textbf{Note:} The red box highlights areas where the correspondence results are significantly not correct.}
 \label{fig:geoca}
\end{figure*}

\subsection{Geometric Calibration}
In order to demonstrate the versatility of our procam registration, we also applied it in projection geometric calibration task. This validates the effectiveness and advantages of the proposed work in real-world scenarios.

Geometric calibration is a critical process that involves precisely determining the geometric parameters of an imaging system, such as a camera or projector. This process is essential for correcting distortions, ensuring accurate alignment, and achieving faithful representation of captured or projected images. In our approach, we leverage projection images generated by our novel text-to-image generation method. These images, when combined with our procam registration model specifically designed for calibration purposes, enable our method to be highly effective in geometric calibration tasks. As illustrated in Fig. \ref{fig:geoca}, our approach demonstrates its efficacy in this domain. Additionally, we conduct a visual performance comparison between the generated images from our method and those sourced from Google Images. Moreover, we evaluate the effectiveness of our proposed deep registration model against previously established pre-trained registration models, further highlighting the robustness and accuracy of our approach.


\begin{figure*}[tb]
 \centering 
 \includegraphics[width=\textwidth]{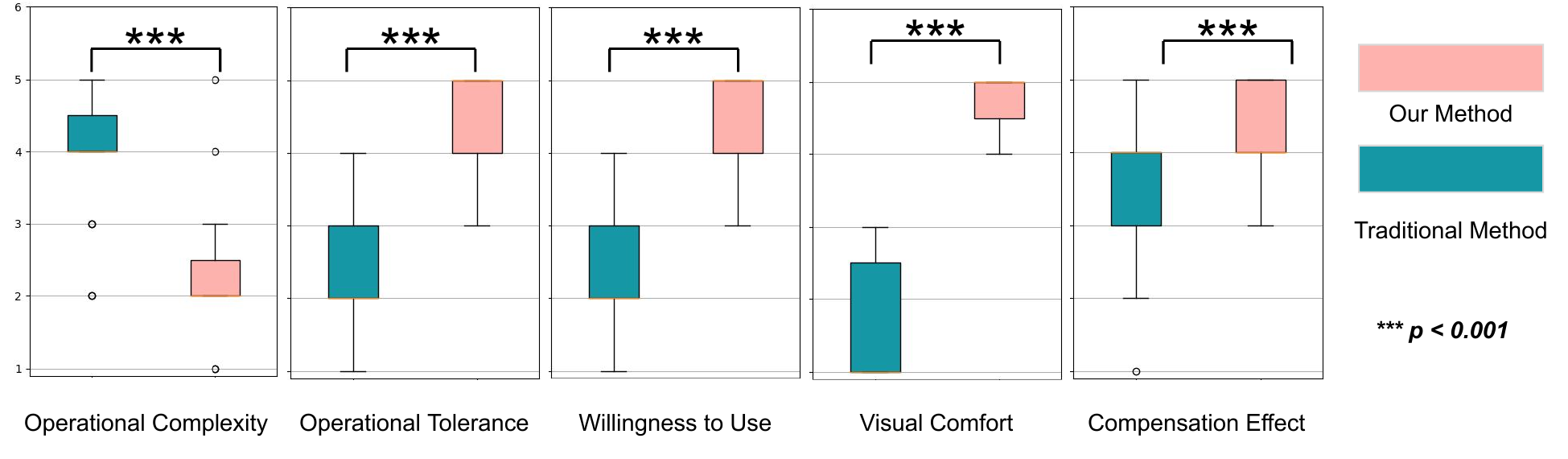}
 \vspace{-2.5em}
 \caption{Results of paired t-test comparing user ratings between proposed and traditional structured light methods. Our method showed significantly lower operational complexity, higher operational tolerance, higher willingness to use, better visual friendliness, and superior compensation effect. Significance levels: *** p<0.001.}
\label{fig:user}
\end{figure*}

\subsection{User Study}

We conducted a user study to evaluate the user experience of traditional geometric calibration (structured light patterns) methods compared to the method proposed in this study for projection geometric correction applications. The experiment involved 19 participants aged between 22 and 27 years (mean: 24.79, standard deviation: 2.06).

As shown in the setup(see Fig. \ref{fig:teaser} (a)), a Marvel movie was projected onto a curved screen, with the projection pre-calibrated geometrically to ensure that participants could see undistorted video frames from their perspective. At random moments, either the projector or the curved screen was moved, causing the projected image to become distorted. We then paused the movie and applied the calibration method to correct the geometry before continuing with the subsequent video frames. After experiencing the display process using both geometric correction methods (traditional structured light and ours), participants were asked to complete a questionnaire (see supplementary materials) to assess the differences between the two methods in terms of operational complexity, tolerance, willingness to use, visual comfort, and compensation effectiveness. 

Figure \ref{fig:user} compares user evaluations of our proposed method and the traditional method across five dimensions using paired t-tests. Our method significantly outperformed the traditional method in all measured aspects. The differences in operational tolerance, willingness to use, and visual comfort were particularly pronounced (p < 0.001), while operational complexity and compensation effectiveness also showed significant improvement (p < 0.01). Scores across all aspects were high, indicating that our method excels in both user experience and technical performance.

\section{Limitation and Future Work}

At present, our method is constrained to performing procam alignment on smooth surfaces due to the underlying smoothness assumptions inherent in our alignment estimation techniques. 
Consequently, the synthetic dataset and the network architecture employed in our approach are insufficient for capturing the full range of surface variations encountered in real-world applications. Specifically, the dataset used for training our deep alignment model was synthesized using image-based techniques with a simplistic model, which does not accurately represent the intricate projection and imaging processes involved in procams systems. As a result, the model trained on this dataset exhibits poor generalization performance when applied to more complex, real-world scenarios. To address these limitations, future work should synthesize a higher-quality dataset using physics-based rendering techniques. Such an approach would enable a more realistic simulation of the projection and imaging processes, accounting for factors such as lens distortion, lighting, surface texture, and geometric complexity. By employing physics-based rendering, the dataset would more accurately reflect the conditions encountered during real-world procam alignment tasks, ultimately improving the model's robustness and generalization capabilities.

In addition, another limitation of our current approach lies in the separation of image generation and deep registration methods. These two processes are currently treated as independent tasks, which leads to suboptimal outcomes in both image generation and pixel correspondence prediction. The two tasks are highly correlated, but we do not leverage the potential relationships that could improve the performance of both methods. Therefore, future work should explore joint optimization techniques that integrate projected image generation with deep registration, allowing both processes to improve one another.

Next, our current method generates a single non-patterned image during alignment, without considering the geometry and texture of the scene. This approach does not fully exploit the potential for optimizing alignment accuracy. A more adaptive approach, where multiple images are generated progressively based on the specific characteristics of the scene, could lead to significant improvements in alignment performance. By dynamically adjusting the projected images to account for the scene's geometry, texture, and other factors, the system could achieve more precise alignments, particularly in challenging environments with complex surfaces.

Finally, while our method demonstrates some ability to handle surface texture during Procams registration, it does not address the radiometric compensation challenges that arise from varying surface reflectivity and lighting conditions. In practice, these radiometric variations can significantly affect the display quality. Therefore, a crucial direction for future research is the development of projector image generation techniques for both geometric and radiometric compensation. By addressing these factors, future approaches could ensure that the projected image not only aligns accurately with the scene’s geometry but also maintains consistent visual quality across different surface textures and lighting conditions. We also would like to extend the technique to multiple projector display applications.

\section{Conclusion}
In this paper, we introduced the first text prompted natural image generation method specifically  targeted for procam registration, offering enhanced design flexibility and a more natural user experience for spatial augmented reality applications. By leveraging local feature descriptors as control conditions, our technique generates images rich in spatial features, greatly improving their suitability for procams registration.

To further enhance accuracy, we also developed a deep neural network to predict projector-camera image correspondences. This approach not only supports precise procam registration but also expands the potential for creating visually meaningful and spatially rich content. Additionally, the method performs well in geometric calibration, ensuring perfect alignment of the projected image with the physical scene. Lastly, the proposed method has also proved to be effective in 3D reconstruction. This approach opens up possibilities for product design and advertising displays based on projector-camera systems.

\begin{acks}
To Robert, for the bagels and explaining CMYK and color spaces.
\end{acks}

\bibliographystyle{ACM-Reference-Format}
\bibliography{template}
\end{document}